# Potential Applications of Artificial Intelligence and Machine Learning in Radiochemistry and Radiochemical Engineering

E. William Webb, PhD, Peter J.H. Scott, PhD
Department of Radiology, University of Michigan, Ann Arbor MI 48109, USA
E-mail: pjhscott@umich.edu

*Introduction*

Radiochemistry for positron emission tomography (PET) applications is a complex amalgam of different areas of expertise. The field combines fundamental organic chemistry and analytical sciences, all under the constraint of timely production for short-lived isotopes ($^{11}$C, $^{18}$F, and $^{68}$Ga) to meet medical demand with sufficient activity and purity. Taken as a whole, these constraints have barred all but a few small molecules from being studied in animals and/or being commercialized. While the generation of novel molecules for further study is addressed elsewhere in this issue, the role of radiochemists in the radiotracer pipeline (Scheme 1) is to identify which site in the molecule is best for labeling, to determine what is the ideal strategy for labeling at that site, to optimize the chemistry to effectively produce the radiolabeled product compound, and finally to develop an appropriate analytical technique to verify the identity and purity of the labeled molecule. To date, the main approach to achieving these ends has been through substantial trial and error, consuming a great deal of time (both human and instrument) and resources. As many of the tools used in artificial intelligence and machine learning become accessible to researchers, there is mounting potential to turn these tools to the problems encountered in the production of radiolabeled molecules for PET applications.[1] While a commonly bandied "buzzword" meant to evoke a superhuman ability to understand a system, artificial intelligence is simply the "intelligence" displayed by machines that emulates the "natural intelligence" of animals or humans through the application of mathematical and computer science algorithms for evaluating data ("machine learning") and executing decisions. These do not replace humans in the scientific process; rather, these can be thought of as convenient "experts" and tools to complement and enhance chemists in the field. In this perspective we outline some of the potential applications of artificial intelligence in the field of radiochemistry.

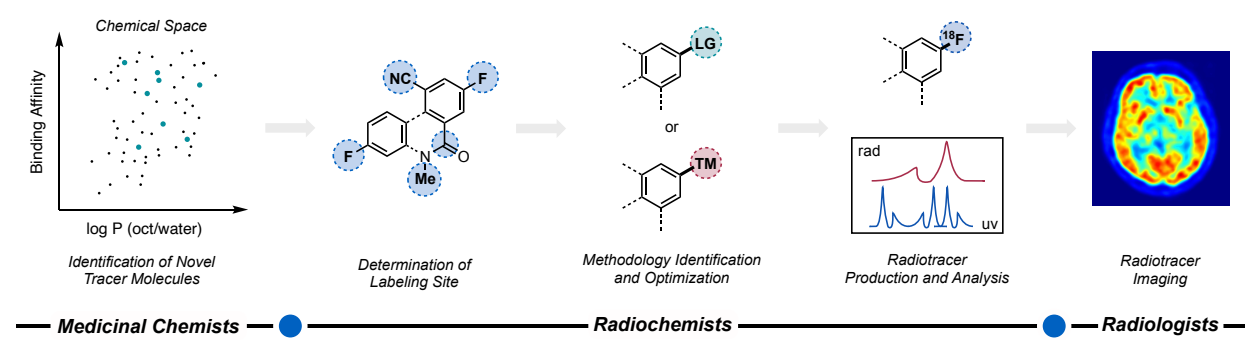

*Scheme 1*. Schematic of radiotracer development and production workflow and areas of radiochemist involvement.

*How Machine Learning Works*

There is some disagreement as to whether Machine Learning (hereafter ML) is a subfield of artificial intelligence (AI) or a separate field that overlaps with some of the AI field.[2,3] Regardless of this disagreement, a functional definition is that AI is a non-biological system that displays human-like intelligence through rules while ML is represented by algorithms that learn from data and examples.[3] An AI could be developed by an expert to run through a set of encoded decisions (a series of if-then statements), similar to an expert's logical and experiential workflow. Following the encoding, an AI agent would then be capable of following the same logic for a novel input to determine a predicted output just as the expert would provide based on their logic and experience. Alternatively, ML could be used to develop a similar set of encoded decisions starting from data or examples using a variety of algorithms, just as the expert once learned.[3] The focus of this perspective will be on the development of artificial intelligences in radiochemistry from data, without the intermediacy of an expert, and thus is most aptly described as machine learning for radiochemistry and radiochemical engineering.

ML has been traditionally broken into three categories: (1) supervised learning, in which the algorithms are presented with data containing example inputs (features) and corresponding desired outputs (labels); (2) unsupervised learning, in which the labels are not given, and algorithms discover underlying structure; and (3) reinforcement learning, in which an algorithm interacts with an environment and is provided with rewards to maximize or losses (penalties) to minimize.[3,4] Depending on the question under study by the researcher, the appropriate mode of machine learning may be different. For example, examining a large set of chest x-ray images using unsupervised learning techniques may identify certain characteristics consistent with a pathology (perhaps an increase in localized densities or increased heart size) without knowing those attributes were a part of the diagnosis. Alternatively, x-ray images labeled with a diagnosis may be used to train a supervised model to diagnose on the basis of an image. However, for the purposes of radiochemical engineering and radiochemistry, supervised and reinforcement learning methods are the most easily applicable.

Whatever the approach and the problem under study, care must be taken to verify that the model generalizes.[4] With small datasets and a large number of features, overfitting can readily occur in supervised learning.[4,5] This functionally tests whether the model can "memorize" rather than correctly perform the desired function. In unsupervised learning contexts, clustering and attribute features can be identified simply due to the original dataset. Testing an additional dataset composed of similar data to verify that the same features are recognized prevents making assertions that will not hold.[5,6]

*Identification of Optimal Site and Strategy for Labeling*

After identifying an appropriate target molecule, the key part of radiochemistry is to distinguish which site is optimal for labeling from a metabolic stability perspective and radiochemical accessibility.[7] Automated retrosynthetic analysis dates back to proposals by Corey in the late 1960s[8,9], and as accessibility has risen with hardware capabilities, additional implementations of machine learning and artificial intelligence as applied to retrosynthetic analysis has seen an upswing.[10] Schematically, older versions of the programs developed for automated retrosynthetic analysis sought to "disconnect" molecules according to template reactions that were encoded by an expert.[10,11] For example: in MPPF an amide bond may be retrosynthetically broken down into an acyl chloride and an amine according to one template, while another template breaks down the molecule into an amide and an electrophile, while still another may break the same amide bond into a palladium-catalyzed carbonylation-amination (Figure 1A). The program iteratively applies these templates to the building blocks produced by each disconnection until arriving at a set of building block molecules that are commercially available or unable to be disconnected into simpler species using the set of templates. This produces an entire "tree" of routes that converge upon the target molecule (Figure 1B). However, this "tree" contains some effective routes and some ineffective routes.[10] To identify the most viable path for synthetic efforts, the various routes need to be ranked by some criteria. The various retrosynthetic routes may be scored by "greenness", by length of route, commercial availability of building block materials, or some other criterion developed alongside the program.[10–12] More recent advances have used machine learning and reaction databases to eliminate the need for an expert in the construction of template sets and more sophisticated scoring systems that measure the feasibility of the forward reactions to determine the viability of a given retrosynthetic route.[10]

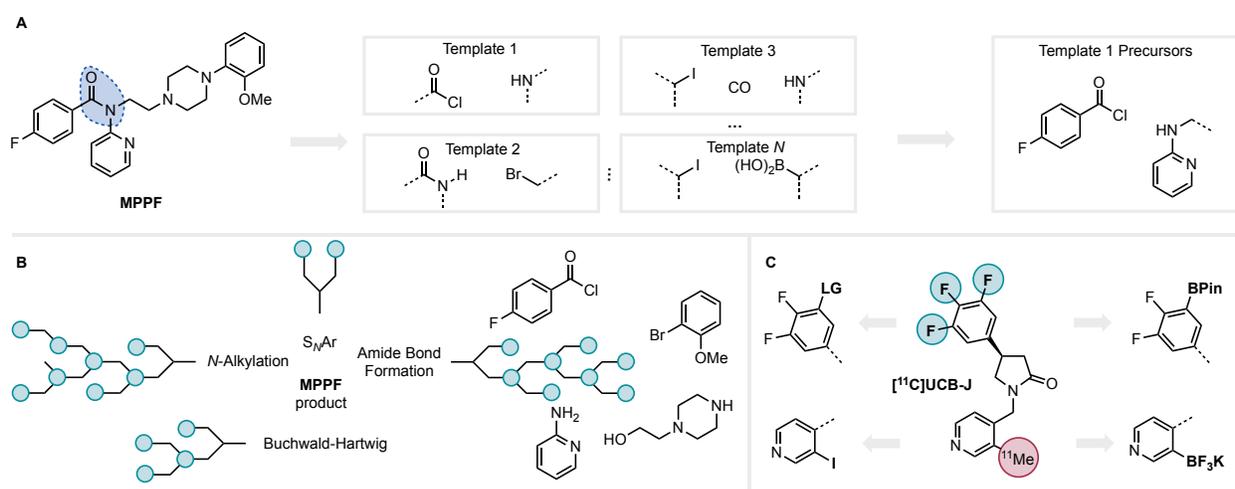

*Fig 1.* (a) Template based retrosynthetic analysis of a complex molecule; (b) Exhaustive deconstruction and generation of multiple potentially viable routes; (c) Retro-radio-synthetic analysis of [¹¹C]UCB-J.

In contrast to typical, multi-step organic synthesis, radiochemists are specifically focused on just one step: the incorporation of the radioisotope, ideally as the last step in the synthetic pathway. This last step is further complicated as different radiolabeling strategies may require completely different starting materials, conditions, workup, or purification strategies, all of which must be completed while limiting radiation exposure, using standardized equipment, and incorporating sufficient activity for transport to the scanning suite and completion of the imaging study. For example: apply a retro-*radio*-synthetic approach to [¹¹C]UCB-J as a prospective compound for labeling (Figure 1C).[13] Any of the aryl fluorides (highlighted in green) could potentially undergo effective labeling via $S_NAr$[14] or a transition metal-mediated radiofluorination with [¹⁸F]fluoride.[15] Alternatively at another site, an iodopyridyl moiety could be labeled with [¹¹C]MeLi[16] while the employed strategy of methylation with [¹¹C]MeI of a pyridyl trifluoroborate[13] offers still another labeling chemistry (highlighted in red). *Ab initio*, the selection of which strategy to pursue is an extremely difficult problem. An $S_NAr$ approach would necessitate the formation of any of several highly reactive electrophiles, while using a transition-metal mediated approach, such as copper-mediated radiofluorination, may not tolerate *ortho*-fluorine substitution.[14,15] Taken altogether, this obligates either a brute force approach to radiosynthesis, testing all possible methodologies presented in the literature and variations thereof in a limited throughput manner or testing the most easily accessed methodology, which, even if modestly successful, may not be optimal. If either of those approaches fail to provide sufficient activity in a timely fashion, all too often the target molecule is discarded as "unlabelable."

Just as machine learning provides additional tools for retrosynthetic analysis, machine learning has potential as a fundamental tool for constructing a retro-radio-synthesis tool. Similar to traditional retrosynthetic analysis tools, template reactions for radiolabeling can be developed. However, the difference between traditional retrosynthetic tools and any potential tool for retro-radio-synthesis is in the scoring function defined for radiochemistry. Any program for this purpose will seek to maximize feasibility, activity, and specific activity while minimizing time for the all-important radiolabeling step and metabolic breakdown[7,17,18] rather than the economics or "greenness" of the route. Foundationally, this will require a change in the way radiochemistry methodology development is conducted. Methodologies will need to be conducted with the intent of translation of the corresponding dataset into a machine learning model that can be further augmented as additional methodologies are developed.[10] For this to occur, substrate scopes, the main datasets of methodologies, must be redefined. This redefinition of substrate scope evaluations must:

(1) Include not just successful reactions but unsuccessful reactions;
(2) Substrates that span the chemical space of accessible *or* pertinent substrates;
(3) "Clean" data.

Only by following these constraints can robust ML models be developed to define the feasibility of any proposed reaction. Reaction chemical space is complex and while two substrates may "seem" similar to a radiochemist, they may perform radically different for a given methodology. As demonstrated by Taylor, *et. al.*,[19] two seemingly similar aryl boronic acid pinacol esters undergo labeling with extremely different efficacy (Figure 2A). Without actively conducting this experiment, it would be difficult to predict this effect as humans would identify these two substrates as near in chemical space. In contrast, with appropriate featurization, most ML algorithms could identify whether these two substrates are neighbors in feature space or not. Given a novel substrate that could potentially be reactive, unreactive model substrates may be nearer in *n*-dimensional chemical feature space than successful scope substrates (Figure 2B). This would lead to the prediction that the novel substrate is more likely to perform similarly to poor performing substrates but without substrate scopes that contain these unsuccessful substrates, such a comparison is unable to be made by models. Unfortunately, the inclusion of ineffective substrates in the literature remains a rarity, except in the case of clearly instructive examples.

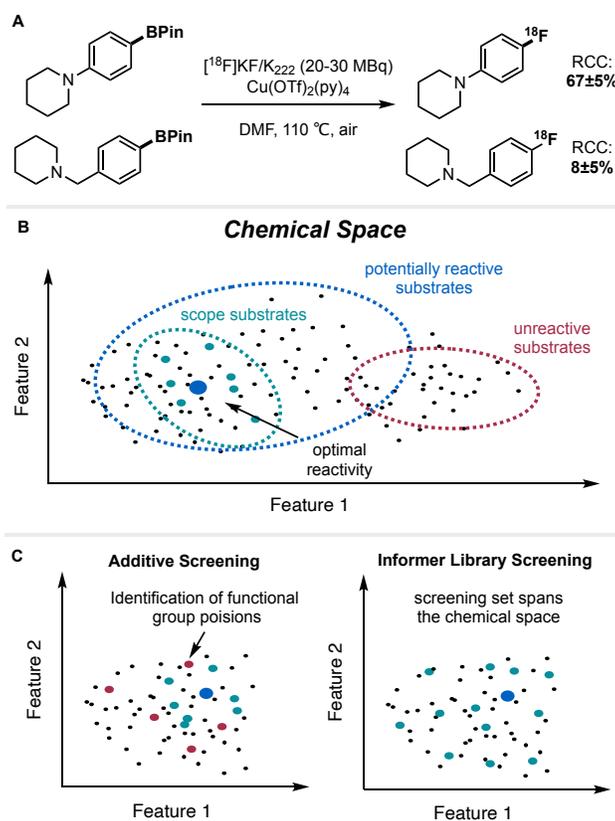

*Fig 2*. (a) Example similar substrates that display drastically different activity[19]; (b) Literature bias of methodology scope towards highly reactive substrates around optimal reactivity rather than potentially reactive or unreactive substrates; (c) Comparison of how different screening strategies span chemical space.

Unsuccessful substrates help define the radiolabeling reaction efficiency surface (Figure 2B). To best define that surface, substrates would be selected to fully span and represent the whole of chemical space. However, spanning chemical space is a daunting and, in truth, impossible prospect. The number of molecules theoretically accessible in small molecule chemical space is over $10^{60}$ molecules.[20] Even when the set of methodology input molecules is reduced to only those functionally pertinent, such as aryl boronate derivatives for a Chan-Lam coupling[21], and then reducing further to the set of commercially available (as an approximate estimation of actual accessibility) brings one into a regime (~$10^6$ molecules). This is still beyond the synthetic accessibility of most labs that lack high throughput experimentation equipment. Because this scale remains synthetically out of reach to most labs, two approaches have become commonplace: additive screening with a model reaction[22] and identification of representative sets[23,24] (Figure 2C). Additive screening is able to readily identify functional groups that act as poisons using a single analytical method but fail to capture the effect those functional groups may have when more proximate to the reactive center.[19,22,23] For the identification of representative libraries, after property calculations are performed on a large number of potential molecules, the set may be reduced through principal components analysis[23] or via finding the most diverse small set using the Kennard-Stone Algorithm.[25,26] These informer sets may not include all potential functionality and may also be combined with additive screening to provide an information-rich dataset for model construction.[24] To date, a standardized set of representative functional group molecules for additive screening has not been demonstrated, but the introduction of a common set would provide a better evaluation of differences between literature methodologies. By attempting to span a wider range of chemical space algorithmically, ML models will demonstrate higher generalizability to novel molecules, in particular for the substrates of labeling interest.

"Clean" data is, unfortunately, the most ambiguous and potentially most important part for the construction of ML datasets for predicting labeling efficiency. There is very little standardized and tabulated data of radiochemical reactions and conditions that is fit to be parsed and mined for AI development. Even the wider organic chemistry field notes the absence of tabulated reaction data.[10] Minute variations in multiple variables that are not clearly annotated in the literature can combine to produce a very large "hidden" variable. Differences in amount of precursor, amount of activity used, the preparation of that activity, or even amounts of different counter ions all may have an effect on labeling efficiency. In a supervised learning problem when two datasets that share a common point are combined, without clear annotation of these differences, one input may map to two outputs in the training set. This will increase the error rate of the model simply due to this "hidden" difference. To achieve so-called "clean data", as many variables as possible need to be annotated and consistent across all experiments. With datasets designed for machine learning, the major hurdle for implementation and use of machine learning can be overcome and the potential of artificial intelligence for radiolabeling chemistry realized.

*Reaction Optimization*

After determining an appropriate target and labeling strategy, radiochemistry becomes an optimization problem, both in reaction development and analytical chemistry. For optimization of a specific reaction, chemist-led or design-of-experiment (DOE) approaches have been most prevalently applied.[27] Fully automated systems have the brute-force capability to conduct many more experiments than a chemist, however, a chemist's intuition may be more efficient in identifying the best experiment to run for optimization.[28,29] While, a chemist's instinct and flexibility may be invaluable, it cannot be parallelized and is dependent on the chemist who, while perhaps an expert on one type of reaction, may not be an expert on the specifically needed reaction. In the case where the chemist is not an expert, a large number of variable conditions may be readily identified from the literature, leading to an exponentially increasing number of potential experiments ("the curse of dimensionality").

As a complement to chemical intuition and to speed navigation of this high dimensional optimization problem, efforts have been undertaken to automate the decision-making process of a chemist so that an optimal set of conditions can be identified in a minimal number of experiments. These algorithms are most similar to a reinforcement learning approach: an initial environment is defined, and on the basis of the outcome of those experiments, additional experiments are selected to maximize (optimize) a reward function (Scheme 2).

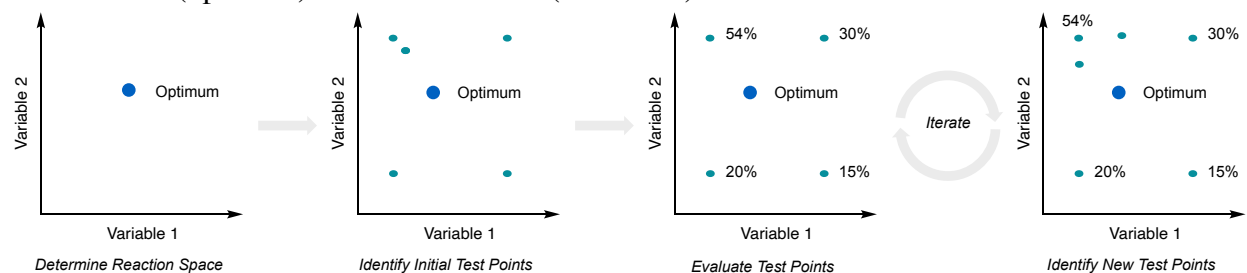

Scheme 2. Schematic for algorithmic and automated optimization.

For a continuous, single objective variable system optimization – for example, identifying optimal amounts of reactants or chromatography gradients – various algorithms have been demonstrated such as the Nelder-Mead simplex method, Stable Noisy Optimization by Branch and Fit (SNOBFIT), and gradient descent optimization.[29,30] However, chemistry is rife with categorical variables as well as continuous variables. For these, alternative approaches like mixed integer linear programming (MINLP) as demonstrated by Jensen and coworkers[28–31], Bayesian optimization[32,33], or Deep Reinforcement Optimization (DRO)[34] may be used to optimize across mixed variables. Even multiple objective optimizations have demonstrated to be feasible.[33]

Taken as a whole, the opportunities for optimization algorithms in radiochemistry are plentiful. Ideal analytical methods that most efficiently and effectively characterize the target molecule can be found in an automated fashion; optimization of reaction conditions can be treated as a multi-objective maximization problem for yield, purity, molar activity, and ease of purification. The high degree of automation currently employed in radiochemistry will facilitate the implementation of these techniques. However, further efforts will be needed to adapt fully automated systems (both hardware and software) into a radiochemical optimization workflow. At present, standard automated synthesis boxes that are commercially available (both cassette-based modules and fixed tube systems)[35] balance current Good Manufacturing Practice (cGMP) features and flexibility for production of various different radiotracers, but are poorly adapted to sequential, fully automated testing. With appropriate equipment and software application programming interfaces (APIs), implementation of artificial intelligence into the workflow for radiochemistry methodology and tracer development becomes accessible to general radiochemistry labs.

*Production and other Concerns*

The workflow of radiochemistry extends past identification and development of a tracer to long-term, repeated production.[36] In the course of long-term production, additional issues may arise that impact effective production. These range from maintenance of automated equipment like cyclotrons and synthesis boxes to changes in the quality of production reagents. Provided a problem in these areas can be defined and reduced to several input factors (for example, electrical current use or coolant temperature may provide alarms for cyclotron maintenance) or sufficient tabulated data collected for unsupervised-learning and identification of underlying patterns, the potential for implementing machine learning in other areas may be readily realized.

*Conclusion*

Radiochemistry, as a field, has readily embraced automated technology to solve various problems including radiation safety and cGMP compliance. For both radiochemistry and radiochemical engineering, machine learning and artificial intelligence offer an additional, powerful tool for evaluating data. This tool will be applied with increasing regularity and offers many time and cost-saving advantages over traditional resource-intensive laboratory approaches. This perspective sheds light on the potential problems to which machine learning may be applied, and how to begin approaching those problems. This chapter is far from exhaustive and the only limit as to what problems are fit for machine learning and artificial intelligence is how well scientists and engineers can define their problems so as to apply these techniques.

*Acknowledgments*

This work was supported by the NIH (Award Number R01EB021155). We also thank Prof. Melanie Sanford and her group members for helpful discussions.

*Disclosure*

The authors declare that they have no conflicts of interest relating to the subject matter of the present review.

*Additional Resources*

Given the rapidly developing nature of artificial intelligence and machine learning, references rapidly become outdated. There are several readily accessible manuals for beginning to construct the coding framework for ML models from O'Reilly. These should remain effective provided the program language in vogue (presently Python, R, or Matlab) does not update. For further concepts and updated applied research on chemistry and AI, the websites of the following professors will likely provide the most up-to-date information: Prof. Alán Aspuru-Guzik, Prof. Connor W. Coley, Prof. Klavs F. Jensen, Prof. Abigail G. Doyle, Prof. Leroy Cronin, Prof. Timothy A. Cernak, and Prof. Scott E. Denmark.